\pdfoutput=1
\documentclass[11pt]{article}

\usepackage[preprint]{acl}

\usepackage{times}
\usepackage{latexsym}

\usepackage[T1]{fontenc}
\usepackage[utf8]{inputenc}

\usepackage{microtype}

\usepackage{inconsolata}

\usepackage{amsmath}
\usepackage{CJKutf8}

\usepackage{booktabs}
\usepackage{multirow}
\usepackage{multicol}
\usepackage{graphicx}
\usepackage{amsmath}
\usepackage{arydshln}
\usepackage{colortbl}
\usepackage{array}

\newcommand{\astfootnote}[1]{
    \let\oldthefootnote=\thefootnote
    \setcounter{footnote}{1}
    \renewcommand{\thefootnote}{\fnsymbol{footnote}}
    \footnotetext{#1}
    \let\thefootnote=\oldthefootnote
}

\title{Analysis of Multi-Source Language Training in Cross-Lingual Transfer}

\author{
    Seong Hoon Lim$^\dagger$, $\;\,$Taejun Yun$^\dagger$, $\;\,$Jinhyeon Kim$^\dagger$, $\;\,$Jihun Choi$^{\ddagger}$, $\;\,$Taeuk Kim$^{*\dagger}$\\
    $^\dagger$Hanyang University, $^{\ddagger}$Sony AI \\
    \small{\texttt{\{dmammfl,tj1616,kimjinhye0n,kimtaeuk\}@hanyang.ac.kr, jihun.a.choi@sony.com}}
}

\begin{document}
\maketitle
\begin{abstract}
The successful adaptation of multilingual language models (LMs) to a specific language-task pair critically depends on the availability of data tailored for that condition. 
While cross-lingual transfer (XLT) methods have contributed to addressing this data scarcity problem, there still exists ongoing debate about the mechanisms behind their effectiveness. 
In this work, we focus on one of the promising assumptions about the inner workings of XLT, that it encourages multilingual LMs to place greater emphasis on language-agnostic or task-specific features.
We test this hypothesis by examining how the patterns of XLT change with a varying number of source languages involved in the process. 
Our experimental findings show that the use of multiple source languages in XLT---a technique we term Multi-Source Language Training (MSLT)---leads to increased mingling of embedding spaces for different languages, supporting the claim that XLT benefits from making use of language-independent information. 
On the other hand, we discover that using an arbitrary combination of source languages does not always guarantee better performance.
We suggest simple heuristics for identifying effective language combinations for MSLT and empirically prove its effectiveness.
\end{abstract}

\astfootnote{Corresponding author.}

\section{Introduction}

\begin{figure}[t]
    \centerline{\includegraphics[width=\columnwidth]{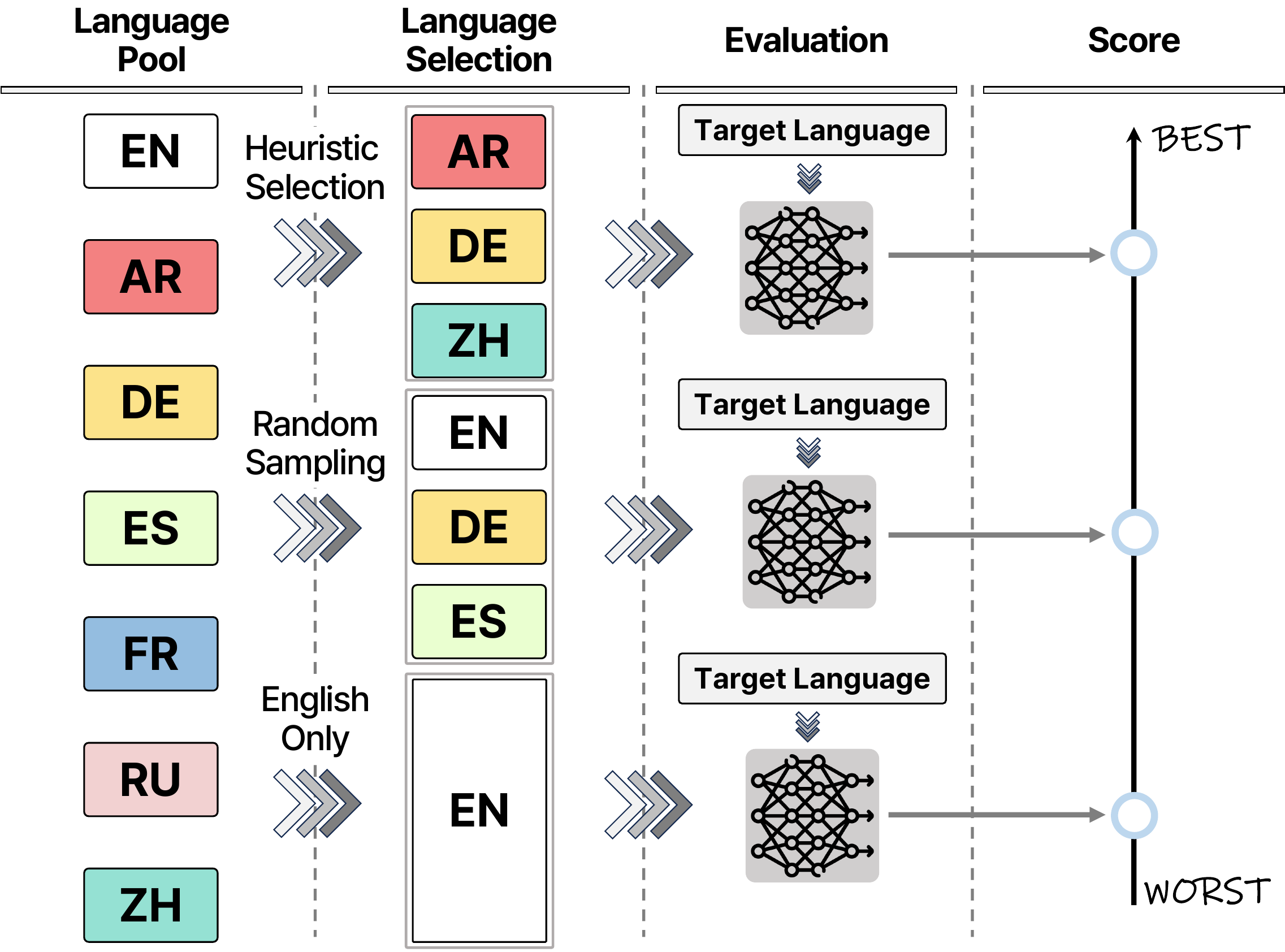}}
    \caption{Overview of the effectiveness of Multi-Source Language Training (MSLT) in cross-lingual transfer. 
    As we adopt more sophisticated approaches for MSLT, we can expect improved performance (from bottom to top).}
    \label{fig:figure1} 
\end{figure}

There has been recent interest in multilingual language models (LMs), where multiple languages are used in training an LM, due to its capability of accepting multilingual inputs and achieving similar or even better performance compared to monolingual LMs at natural language processing (NLP) tasks.
However, despite all these advantages, fine-tuning multilingual LMs using a specific language and an NLP task hinges on acquiring labeled data tailored to a specific purpose, which is not always possible.

To address this challenge amid linguistic resource disparities, cross-lingual transfer (XLT) approaches have been proposed.
In the context of fine-tuning multilingual LMs, XLT leverages resource-rich source languages in fine-tuning to boost performance on low-resource target languages.

Although there is still active debate regarding the inner workings of XLT, a significant body of previous research indicates that multilingual LMs are capable of separating \textit{language-specific} and \textit{language-agnostic} information from text \cite{muller2021first}, and that XLT favors the enhancement of language-agnostic features while reducing the emphasis on language-specific ones (\citealp{qi2022enhancing,tu-etal-2022-prompt,wang2022english}; \textit{inter alia}).

While the standard practice in XLT is to employ only one language as a source for transfer, there exist attempts that have investigated the use of multiple source languages \cite{singh2019xlda,roy2020lareqa,kew2023turning,chai2024xcot,shaham2024multilingual}.
These studies reveal that employing multiple source languages in XLT---the concept we refer to as \textbf{Multi-Source Language Training (MSLT)}---leads to performance improvement.
However, despite it is evident that MSLT has clear advantages, several research questions remain unexplored, including: (1) the specific changes that it induces in the internal states of LMs, (2) the conditions under which it achieves its optimal performance, and (3) the breadth of applications where it can be effectively applied, among others.

In this work, we introduce a series of comprehensive analyses exploring the inner workings of MSLT, along with offering guidance for its effective utilization.
Specifically, we first attempt to illustrate the utility of MSLT with a couple of intuitive visualizations, to clarify the underlying reasons for its effectiveness.
In addition, our experimental results discover various intriguing findings.
Regarding the number of languages involved in MSLT, we observe its positive correlation with performance in general, although the trend tends to plateau as the number of languages exceeds a certain threshold.

Our empirical discovery also suggests that the enhanced diversity of source languages in MSLT does not invariably lead to improved performance. 
This underscores the importance of carefully selecting an effective combination of source languages (see Figure \ref{fig:figure1} for an intuitive illustration).
Therefore, we examine various heuristics that can identify a suitable selection of source languages from an exponentially large pool of possible combinations.
Based on our tests, we propose several competent strategies, some of which are derived from the statistical properties of languages in terms of pretraining, while others utilize linguistic characteristics of languages.
Lastly, we discuss intriguing patterns evident in the group of languages selected as sources for transfer, providing insights into the interplay of languages in MSLT.

\definecolor{green}{HTML}{1AC938}

\section{Related Work}

\definecolor{pink}{HTML}{F38181}
\definecolor{yellow}{HTML}{FF9A00}
\definecolor{sky}{HTML}{3EC1D3}

Multilingual LMs (\citealp{conneau2019unsupervised,scao2022bloom}; \textit{inter alia}) are widely recognized for their ability to process inputs from diverse languages in an integrated manner.
They have recently been the subject of extensive research for uncovering their working mechanisms, and this effort has converged on the robust hypothesis that the internal states of multilingual LMs can be categorized into language-sensitive and language-agnostic components \cite{choenni2020does, chang2022geometry, zhao2020inducing,muller2021first}. 

Meanwhile, cross-lingual transfer (XLT) aims to improve the performance of multilingual LMs on specific tasks in languages with limited resources (i.e., target languages), by utilizing the support from resource-rich languages (i.e., source languages).
Previous research on XLT has explored diverse implementations of the concept.
For instance, \citet{yang2022enhancing} attempt to achieve XLT by blending representations from source and target languages, while \citet{zheng2021consistency} introduce data augmentation techniques such as subword sampling and code-switching. 
\citet{wang2022english} propose a contrastive learning framework aimed at reducing the differences in sentence embeddings between languages. Concurrently, \citet{qi2022enhancing} present a prompt-based approach that facilitates information transfer from a source language to prompts composed in the target language.

Note that the majority of existing methods for XLT \cite{libovicky2020language,tiyajamorn2021language,yang2022enhancing}, including the aforementioned studies, adhere to a common theme: they are crafted to facilitate XLT by prompting the enhancement of language-agnostic or universal features within multilingual LMs.
In this work, we focus on a simple strategy that yields a similar effect, namely, introducing multiple source languages in XLT instead of just one as is customary, deliberately steering clear of complex algorithms.

In the literature, there is empirical acknowledgment that increasing the diversity of source languages in XLT can boost its performance.
\citet{singh2019xlda} suggest that pairing sentences written in different languages (for example, one sentence in one language and another in a different language) is beneficial for improving performance on the cross-lingual natural language inference (XNLI; \citealp{conneau-etal-2018-xnli}) task.
They additionally observe that the performance fluctuates based on the languages from which the sentences are sampled.
\citet{roy2020lareqa} demonstrate a similar trend in question-answering tasks.
Further, \citet{kew2023turning} and \citet{shaham2024multilingual} reveal that MSLT is also advantageous for instruction-tuned LMs.

Despite the recent surge in attention towards MSLT, it is important to highlight the lack of a comprehensive analysis or tailored strategies in this area so far.
For instance, \citet{kew2023turning} consider only pre-defined language sets in their experiments, ignoring the potential of discovering the best combination of source languages.
On the other hand, \citet{singh2019xlda} and \citet{shaham2024multilingual} utilize all available languages without specifically seeking to determine the ideal number of languages for the optimal performance of MSLT.
Our work tackles these limitations, by presenting a series of well-supported illustrations that clarify how and why MSLT operates, along with providing guidelines for enhanced utilization of the MSLT framework.

\section{Terminology}
We first clarify the core concepts and their corresponding acronyms in this section.

\paragraph{Cross-Lingual Transfer (XLT)} 
XLT is a technique applicable to multilingual LMs, particularly when obtaining a satisfactory amount of data for the language-task pair of interest is challenging.
This is built on the assumption that fine-tuning a multilingual model for a task $\mathcal{T}$ in a resource-rich language $\mathcal{A}$ can lead to a synergistic improvement in performance for another language $\mathcal{B}$, especially if $\mathcal{B}$ has limited data available for the task $\mathcal{T}$.

In this work, we focus on the zero-shot XLT setting, under the assumption that no suitable data is available for the target language.

\paragraph{Single-Source Language Training (SSLT)} 
SSLT represents an environment for XLT where multilingual LMs are trained on a single language.
From now on, we adopt the notation `$\text{SSLT}(L)$' to signify the setting where the language $L$ is exclusively utilized as a source language in XLT.

\paragraph{Multi-Source Language Training (MSLT)} MSLT refers to an environment in which two or more source languages are considered for XLT. 
We use the notation `$\text{MSLT}(L_1, L_2, \ldots, L_n)$' to denote the setting where MSLT is conducted with languages $L_1, L_2, \ldots, L_n$.

MSLT, when feasible, enables the possibility of leveraging an exponentially higher number of data points by allowing the use of instances from multiple languages.
However, as the purpose of this study is to analyze the impact of language diversity, we decide to keep consistent the total number of data points used in experiments.
For example, if SSLT(\texttt{en}) utilizes 1,000 data points in English, the corresponding MSLT(\texttt{en}, \texttt{es}) incorporates 500 data points each from English and Spanish, respectively.

\section{Verifying the Effectiveness of MSLT}
\label{section:4}
We present a thorough analysis to confirm the effectiveness of MSLT, and additionally provide insights into its inner workings.
We describe the intuition behind how MSLT enhances XLT performance (\S\ref{section:4.2}) and present visualizations related to our intuition (\S\ref{section:4.3}). 
We also provide empirical findings with respect to the ideal number of source languages in MSLT (\S\ref{section:4.4}).
Section \ref{section:4.1} presents the configurations used throughout the section.

\subsection{Experimental Settings}
\label{section:4.1}
To investigate the inner workings of MSLT, we conduct experiments using XLM-RoBERTa$_{\text{Base}}$ (XLM-R$_{\text{Base}}$; \citealp{conneau2019unsupervised}), a compact multilingual encoder that covers around 100 languages.
We select three tasks from the XTREME benchmark \cite{hu2020xtreme} as our testbed for analysis: named entity recognition (WikiANN; \citealp{pan-etal-2017-cross}), cross-lingual natural language inference (XNLI; \citealp{conneau-etal-2018-xnli}), and cross-lingual paraphrase identification (PAWS-X; \citealp{yang-etal-2019-paws}).
More details on the datasets and training processes are specified in Appendix \ref{appendix:dataset} and \ref{appendix:hyperparameters}, respectively.

\begin{figure}[!t]
    \centerline{\includegraphics[width=\columnwidth]{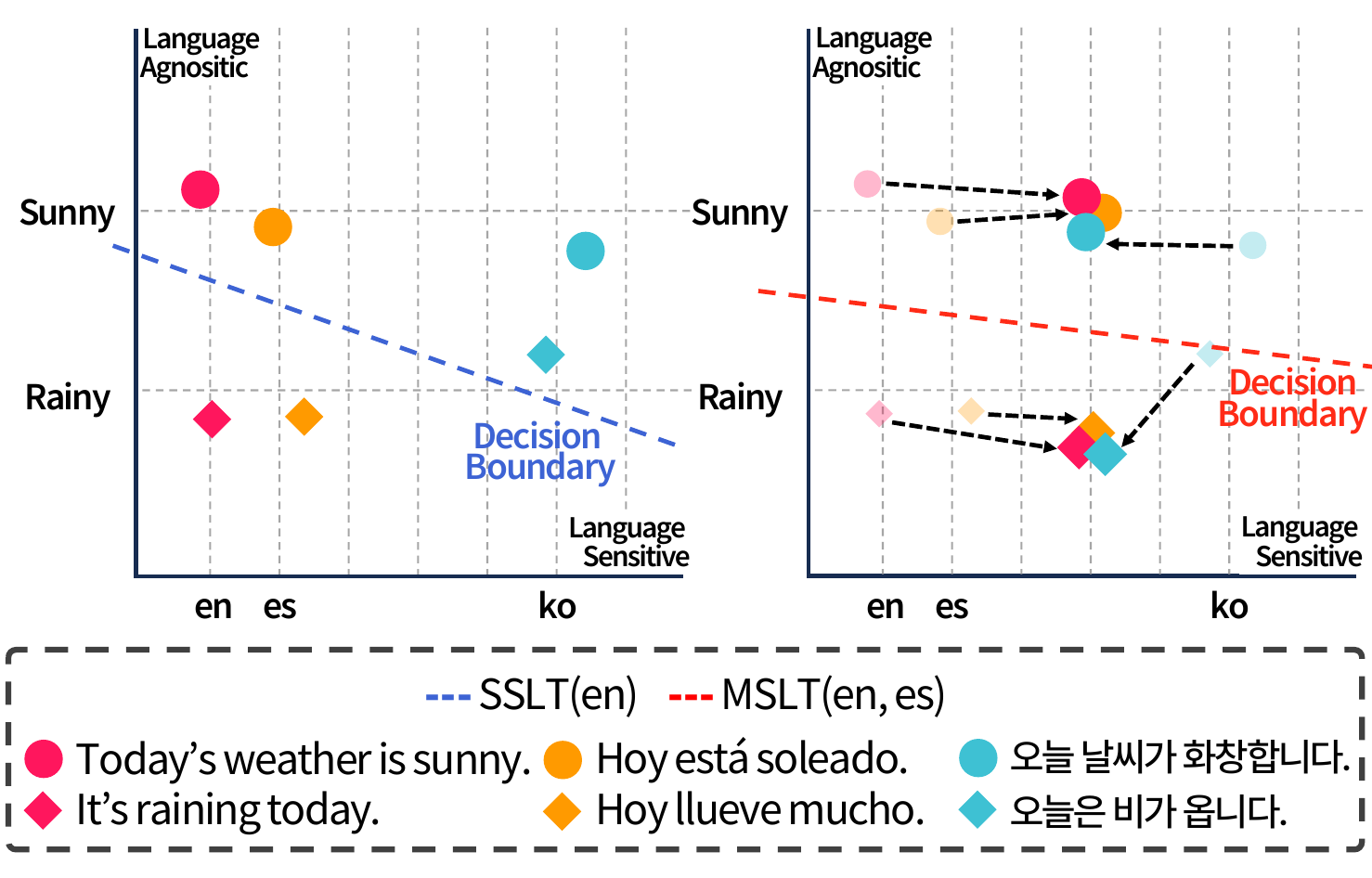}}
    \caption{A conceptual illustration of the advantages of MSLT over SSLT. 
    The left illustrates the training process of an LM using only \textcolor{pink}{English (\texttt{en})} (i.e., $\text{SSLT}(\texttt{en})$), while the right represents MSLT with \textcolor{pink}{English (\texttt{en})} and \textcolor{yellow}{Spanish (\texttt{es})} (i.e., $\text{MSLT}(\texttt{en}, \texttt{es})$). 
    Incorporating more source languages enhances language-agnostic features and blurs language-specific ones, potentially improving effectiveness for unseen languages such as \textcolor{sky}{Korean (\texttt{ko})}.
    }
    \label{fig:figure2} 
\end{figure}

\subsection{Advantages of MSLT over SSLT}
\label{section:4.2}
Rooted in the widely accepted notion within the literature that multilingual LMs can categorize language-specific and language-agnostic concepts \cite{muller2021first}, there is a prevalent belief that XLT can be more effectively activated by encouraging the influence of language-agnostic features while diminishing the role of language-specific ones.
We draw on this wisdom to intuitively explain why MSLT generally outperforms SSLT.

\begin{figure*}
    \centering
    \includegraphics[width=\textwidth]{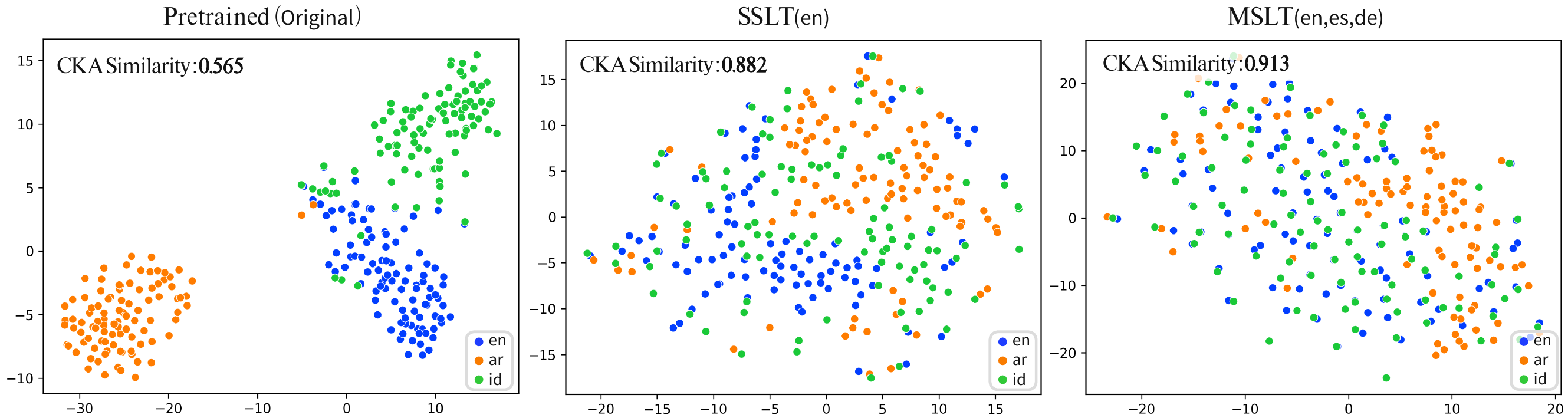}
    \caption{
    Visualization of embeddings and corresponding CKA similarities \cite{kornblith2019similarity} for 3 languages: \textcolor{blue}{English (\texttt{en})}, \textcolor{orange}{Arabic (\texttt{ar})}, and \textcolor{green}{Indonesian (\texttt{id})}. Note that \textcolor{blue}{English} is used in both SSLT \& MSLT, whereas \textcolor{orange}{Arabic} and \textcolor{green}{Indonesian} are not. 
    Therefore, we can observe the impact of SSLT \& MSLT on both languages seen and unseen during training.
    \textbf{Left}: the original XLM-R.
    \textbf{Center}: XLM-R after $\text{SSLT}(\texttt{en})$.
    \textbf{Right}: XLM-R after $\text{MSLT}(\texttt{en}, \texttt{es}, \texttt{de})$.
    We find that while SSLT promotes language-agnostic alignment in the semantic space, MSLT enhances this further, 
    leading to a more integrated space for languages.
    }
    \label{fig:figure3}
\end{figure*}

Figure \ref{fig:figure2} illustrates the potential advantages of MSLT over SSLT.
When data from a single language is used in training for XLT, as in the case of SSLT, it may be challenging to guide a model to establish a decision boundary for classification (indicated by \textcolor{blue}{blue} in Figure \ref{fig:figure2}) that can robustly function across different languages.
Conversely, with MSLT, the model is exposed to signals from a diverse range of languages, encouraging more frequent exploitation of language-agnostic features.
This leads to representations that are comparatively language-independent, enabling the identification of a more robust decision boundary applicable across different languages (\textcolor{red}{red} in Figure \ref{fig:figure2}).
In summary, we speculate that MSLT provides a more advantageous environment for XLT compared to SSLT, by underscoring the utility of language-agnostic features.

\subsection{Visualization of Embeddings after MSLT}
\label{section:4.3}
To better understand the impact of MSLT and empirically validate the hypothesis presented in \S\ref{section:4.2}, we conduct a comparative study by visualizing embeddings before and after applying SSLT and MSLT.
To be specific, we first train the XLM-R$_{\text{Base}}$ model with data instances from the XNLI dataset, resulting in three variants: (1) the original XLM-R, (2) XLM-R with SSLT(\texttt{en}), and (3) XLM-R with MSLT(\texttt{en},\texttt{es},\texttt{de}).\footnote{\texttt{en}: English, \texttt{es}: Spanish, \texttt{de}: German.}
We then visualize the embeddings computed by each model variant,\footnote{Embeddings are derived by mean-pooling the hidden states from the last layer of the language model.} using a set of sentences from the Parallel Universal Dependencies (PUD) treebanks \cite{de2021universal}.
This decision stems from our goal to evaluate the model's general capabilities, not just its performance on fine-tuning tasks, but also its effectiveness with out-of-domain data instances.
We make use of t-SNE \cite{vandermaaten2008tsne} for dimension reduction.

In Figure \ref{fig:figure3}, we observe that the XLM-R model with MSLT (the rightmost) demonstrates significantly better language-agnostic alignment for unseen languages (\texttt{ar} and \texttt{id}), compared to both (1) the original XLM-R and (2) XLM-R after SSLT. This implies that the diversification of source languages prevents a model from learning language-specific features and encourages it to be more language-independent, which is beneficial for achieving better performance in XLT.

\subsection{On the Optimal Number of Languages}
\label{section:4.4}
In the previous part, we qualitatively demonstrated the effectiveness of MSLT through graphical illustrations.
Here, we present a supplementary quantitative analysis on the impact of the number of source languages on XLT, addressing two research questions: (1) Is MSLT consistently superior to SSLT, at least in the evaluated environments? (2) What is the optimal number of languages for MSLT?
We select English, Spanish, German, Chinese, and French as candidates for source languages and evaluate the performance of different language groups on cross-lingual datasets, calibrating their size from 1 (SSLT) to 5 (MSLT with five languages).\footnote{The scores of groups with the same size are averaged.}

\begin{figure}
    \centering
    \includegraphics[width=\columnwidth]{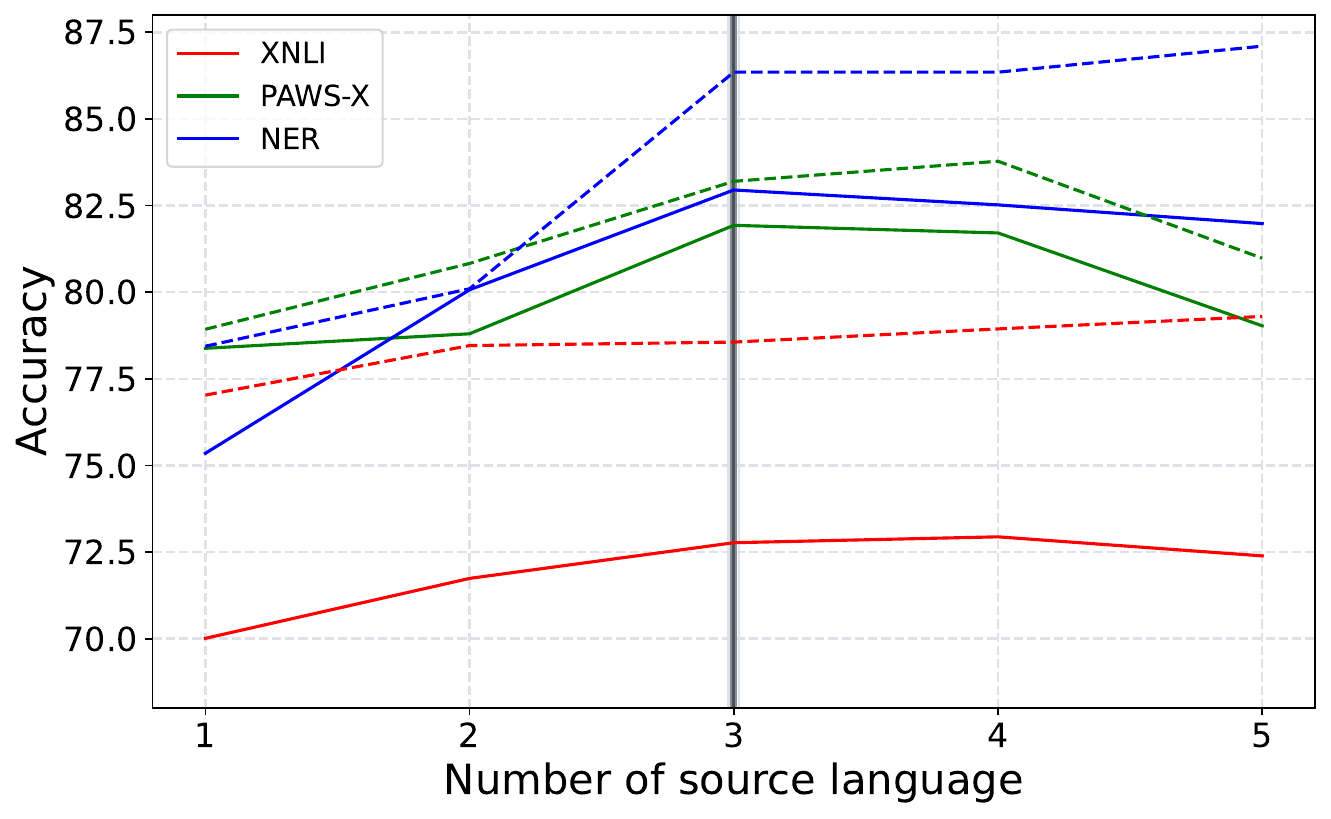}
    \caption{Performance of XLT can vary depending on the number of source languages.
    The solid lines correspond to XLM-R$_{\text{Base}}$ and dotted lines to XLM-R$_\text{Large}$.
    }
    \label{fig:figure4}
\end{figure}

Figure \ref{fig:figure4} reveals that regardless of the task type or data quantity, task performance remarkably improves as the number of source languages increases from 1 to 3, demonstrating the effectiveness of MSLT.
However, when the size of the group exceeds 3, the performance exhibits minimal improvement or even declines in some cases.\footnote{In the case of XLM-R$_\text{Large}$, performance shows a slight improvement even with more than three languages.}
Taking into account the outcomes of our experiments and the expenses involved in creating datasets across various languages, we propose opting for three source languages as a pragmatic and sensible selection for MSLT. 
Consequently, we make use of three-language groups in the forthcoming experiments.

\section{Language Set Composition in MSLT}
\label{section:5}
In the previous section, we demonstrated that MSLT exhibits markedly superior performance on XLT compared to SSLT.
Assuming the selection of $k$ source languages from a total of $n$ languages, there exist $\genfrac(){0pt}{2}{n}{k}$ possible combinations.
Given that $n$ can be significantly larger than $k$, fine-tuning with every possible combination of source languages to identify the optimal mix is not time- and resource-efficient. 
Therefore, it is necessary to select an appropriate combination more efficiently.
In this section, we first establish the importance of selecting an effective set of source languages and then proceed to examine various criteria for choosing an optimal source language combination.

\newcolumntype{A}{>{\centering}p{2cm}}
\newcolumntype{B}{>{\raggedright}p{1.6cm}}
\newcolumntype{D}{>{\raggedright}p{2.2cm}}

\begin{table*}[!t]
    \small
    \centering
    \setlength{\tabcolsep}{3pt}
    \begin{tabular}{B D A A A A A}
        \toprule[1.5pt]
        \multirow{2.5}{*}{\shortstack[c]{\textbf{Method}}}& {} & \multicolumn{2}{c}{\textbf{Encoder}} & \multicolumn{3}{c}{\textbf{Decoder}}
        \\\cmidrule(lr){3-4}\cmidrule(lr){5-7}
        & & WikiAnn & XNLI & XCOPA & XWinograd & XStoryCloze \cr\midrule
   
        MIN & {} & \cellcolor[HTML]{F38181}76.30 (35) & \cellcolor[HTML]{F38181}70.97 (35) & \cellcolor[HTML]{F38181}48.87 (35) & \cellcolor[HTML]{F38181}48.15 (35) & \cellcolor[HTML]{F38181}51.05 (35) \cr

    \midrule
        
        \multicolumn{2}{l}{Size of Pretraining Data} & \cellcolor[HTML]{F7B5B5}78.52 (31)&\cellcolor[HTML]{D8EEFD}72.87 (16) &\cellcolor[HTML]{FCFEFF}50.35 (17)&\cellcolor[HTML]{D7EDFC}54.77 (12)&\cellcolor[HTML]{FDEDEE}54.05 (26) \cr
    
        \multicolumn{2}{l}{Vocab Coverage} & \cellcolor[HTML]{F7B5B5}78.52 (31)&\cellcolor[HTML]{FCFEFF}72.46 (26) &\cellcolor[HTML]{FEFFFF}50.32 (22)&\cellcolor[HTML]{80C6F3}58.21 (2)&\cellcolor[HTML]{FDEDEE}54.05 (25) \cr
    
    \midrule
    
        \multicolumn{2}{l}{Embedding} &\cellcolor[HTML]{A4D6F7}85.58 (4)&\cellcolor[HTML]{FDF1F1}72.26 (31)&\cellcolor[HTML]{F38181}48.85 (34)&\cellcolor[HTML]{FCFEFF}53.33 (19)&\cellcolor[HTML]{A5D7F7}56.98 (7) \cr
        
        Lang2Vec & $\to$ Syntax & \cellcolor[HTML]{A1D5F7}85.69 (3)&\cellcolor[HTML]{89CAF4}73.76 (3) &\cellcolor[HTML]{91CEF5}51.57 (4)&\cellcolor[HTML]{B9E0F9}55.92 (7)&\cellcolor[HTML]{96D0F5}\textbf{57.42} (2) \cr
    
        {} & $\to$ Phonology  & \cellcolor[HTML]{B5DEF8}84.82 (8)&\cellcolor[HTML]{AAD9F7}73.39 (7) &\cellcolor[HTML]{FCDADA}49.88 (27)&\cellcolor[HTML]{80C6F3}\textbf{58.21} (1)&\cellcolor[HTML]{E8F6FD}55.12 (18) \cr
    
        {} & $\to$ Inventory  & \cellcolor[HTML]{98D1F6}\textbf{86.07} (2)&\cellcolor[HTML]{89CAF4}\textbf{73.77} (2) &\cellcolor[HTML]{EEF8FE}50.50 (11)&\cellcolor[HTML]{81C7F4}58.20 (3)&\cellcolor[HTML]{B9E0F9}56.45 (10) \cr
    
        {} & $\to$ Family  & \cellcolor[HTML]{B5DEF8}84.82 (8)&\cellcolor[HTML]{AAD9F7}73.39 (7) &\cellcolor[HTML]{FCDADA}49.88 (27)&\cellcolor[HTML]{80C6F3}\textbf{58.21} (1)&\cellcolor[HTML]{E8F6FD}55.12 (18) \cr
    
        {} & $\to$ Geometry  & \cellcolor[HTML]{A4D6F7}85.58 (4)&\cellcolor[HTML]{B7DFF9}73.24 (9) &\cellcolor[HTML]{84C8F4}\textbf{51.72} (2)&\cellcolor[HTML]{82C8F4}58.12 (6)&\cellcolor[HTML]{B2DDF8}56.62 (9) \cr

    \midrule
    
        MAX & {} & \cellcolor[HTML]{80C6F3}87.05 (1) & \cellcolor[HTML]{80C6F3}73.86 (1) & \cellcolor[HTML]{80C6F3}51.76 (1) & \cellcolor[HTML]{80C6F3}58.21 (1) & \cellcolor[HTML]{80C6F3}57.98 (1) \cr
    
    \bottomrule[1.5pt]
    \end{tabular}
    \caption{Results of the proposed criteria for source language selection, evaluated across five different test datasets.
    The numbers in parentheses indicate the ranking of the specific language set chosen by each method among the total of 35 possible combinations, with higher ranks (indicating better performance) closer to blue and lower ranks (indicating inferior performance) closer to red.
    We confirm that Lang2Vec-based methods are proficient in proposing useful source language sets.
    }
    \label{tab:table1}
\end{table*}

\subsection{Experimental Settings}
We perform experiments utilizing the BLOOM-7B model \cite{scao2022bloom}, in conjunction with the previously utilized XLM-R.
We fix the size of the source language set at 3, and its elements are chosen from a pool of 7 language candidates,\footnote{Arabic (\texttt{ar}), German (\texttt{de}), English (\texttt{en}), Spanish (\texttt{es}), French (\texttt{fr}), Russian (\texttt{ru}), Chinese (\texttt{zh}).} which are commonly considered resource-rich languages.

When applying MSLT to XLM-R, we reuse the WikiAnn and XNLI datasets for fine-tuning and evaluation, as was done in the previous section.\footnote{PAWS-X is omitted owing to its limited linguistic variety.}
For experiments with BLOOM-7B, we leverage separate datasets for instruction-tuning and evaluation, respectively. 
By introducing this instruction-tuning setting, we aim to evaluate whether MSLT is also effective in this new paradigm, which would further extend the applicability of MSLT.
BLOOM-7B is instruction-tuned using the Bactrian-X dataset \cite{li2023bactrianx}. For evaluation, we employ another suite of datasets, which includes XCOPA \cite{ponti-etal-2020-xcopa}, XWinograd \cite{muennighoff-etal-2023-crosslingual}, and XStoryCloze \cite{lin-etal-2022-shot}.
Also note that we fine-tune BLOOM-7B with QLoRA \cite{dettmers2023qlora} for computational efficiency and to ascertain how well MSLT operates in a parameter-efficient fine-tuning setup.
Details on the experimental settings can be again found in Appendix \ref{appendix:dataset} and \ref{appendix:hyperparameters}.

\begin{figure}[t]
    \centering
    \includegraphics[width=\columnwidth]{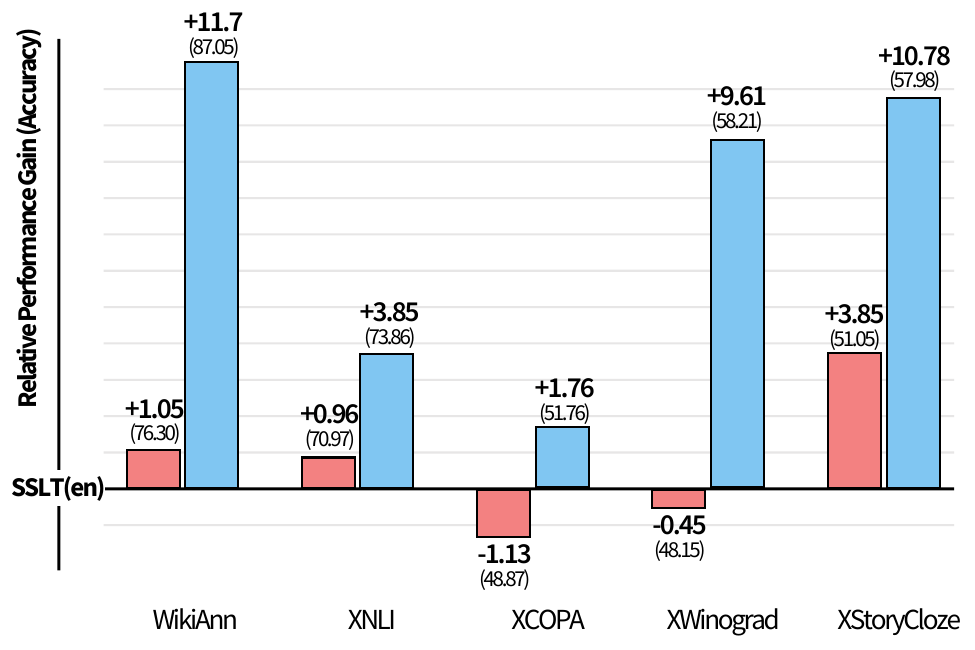}
    \caption{
    Relative performance gaps vary with source language combinations, reaching as high as a 10-point difference between the \textcolor{blue}{best} and \textcolor{red}{worst} options.
    The worst combinations of MSLT in XCOPA and XWinograd even harm performance compared to SSLT, highlighting the need for careful source language selection.
    }
    \label{fig:figure5}
\end{figure}

\subsection{The Harmony of Languages Matters}
\label{section:5.2}
We compare the performance of MSLT across various source language combinations. 
According to our experiments, the magnitude of performance improvement varies greatly depending on the combination used, and in some cases, a poor choice may lead to performance deterioration.
Figure \ref{fig:figure5} represents the upper and lower bounds of the performance of MSLT, along with that of SSLT as a baseline.
A notable performance gap is observed between the optimal and worst combinations in most experiments.
In certain tasks, e.g., XCOPA and XWinograd, MSLT with the least effective combination fails to outperform SSLT.
These findings imply that the combination of source languages in MSLT has a substantial impact on performance, highlighting the need to establish criteria for selecting appropriate source languages.

\subsection{Criteria for Source Language Selection}
\label{section:5.3}
We further explore the criteria for selecting source language combinations by empirically testing several hypotheses related to them.

\subsubsection{Size of Pretraining Data}
\label{sec:5.3.1}
We examine whether using a combination of the most frequent languages in the pretraining data leads to improved performance in MSLT. 
The underlying motivation is that the proportion of a language in the pretraining data might correlate with how much a model acquires the knowledge specific to that language during pretraining. 
Based on the volume of pretraining data for each language, as reported in the original papers \cite{conneau2019unsupervised, scao2022bloom}, we select three languages from the source language pool for fine-tuning: English, Russian, and German for XLM-R, and English, Simplified Chinese, and French for BLOOM-7B.

\subsubsection{Vocabulary Coverage}
\label{sec:5.3.2}
Inspired by the findings of \citet{pires-etal-2019-multilingual}, which suggest that XLT depends on vocabulary overlap between source and target languages, we test a heuristic that chooses source languages with extensive vocabulary coverage. 
Suppose there are $n$ source languages available, represented as $\mathcal{L}=\{L_1, \cdots, L_n\}$, and let the vocabulary set for each language $L_i$ be denoted by $V_{L_i}$.
The coverage of a candidate set $\mathcal{L}_C \subseteq \mathcal{L}$ is then defined by
\[
\text{Coverage}(\mathcal{L}_C) = \frac{\left| \bigcup_{L_i \in \mathcal{L}_C} V_{L_i} \right|}{ \left| \bigcup_{L_i \in \mathcal{L}} V_{L_i} \right| }.
\]
We anticipate that language combinations possessing extensive lexical breadth will encompass a broader spectrum of the target language's vocabulary, ultimately leading to better XLT scores.

\subsubsection{Linguistic Diversity} 
\label{sec:5.3.3}
The last criterion we examine is the selection of source languages that exhibit distinct linguistic properties.
This criterion shares a similar motivation with the previous ones, aiming to use languages that can generalize to other unseen languages. However, it uniquely posits that utilizing as diverse a range of source languages as possible will lead to a more stable and comprehensive modification of the model's internal mechanism.

To achieve this, we first represent each language as language vectors and then identify a candidate set whose language vectors have the \textit{least} similarity to each other.
We introduce Lang2Vec \cite{littell-etal-2017-uriel}, a framework offering a suite of vectors for each language that mirrors its linguistic properties.
In particular, we test five variants of Lang2Vec, i.e., Syntax, Phonology, Inventory, Family, and Geometry vectors.
Formally, given a set of languages $\mathcal{L} = \{ L_1, \ldots, L_n\}$ and their language vectors $\{ v_1, \ldots, v_n \}$, 
the diversity score of a combination $\mathcal{L}_C \subseteq \mathcal{L}$ is defined by the sum of pairwise cosine similarity of language vectors:
\[ \text{Diversity}(\mathcal{L}_C) = \sum_{ \{L_i, L_j\} \subseteq \mathcal{L}_C} \left( 1 - \text{sim}(v_i, v_j)\right). \]
Since our objective is to find a combination $\mathcal{L}_C$ having the maximum diversity, i.e. the minimal uniformity, thus the desired combination is obtained by
${\arg\max}_{\mathcal{L}_C} \text{Diversity}(\mathcal{L}_C) $.

On the other hand, we consider an extra baseline called just `Language Embedding from LMs' or just `Embedding'.
This process is similar to that of Lang2Vec, but the vectors here are derived directly from pretrained LMs, specifically XLM-R or BLOOM-7B, depending on the case.
The language vector for each language is calculated by averaging the embeddings of sentences in that language.

\begin{table}[t!]
    \scriptsize
    \centering
    \begin{tabular}{l c c c c c}
    \toprule[1.2pt]
         \textbf{Method}& WikiAnn & XNLI & XCOPA & XW & XSC \\
    \midrule
        \rowcolor[gray]{0.9}Min&15&15&15&15&15 \\
    \midrule
        PT Data Size&12&6&\textbf{5}&6&11\\
        Vocab Coverage&12&9&10&\textbf{2}&10\\
        Embedding&\textbf{1}&13&9&9&7\\
        L2V-Syn&\textbf{1}&\textbf{2}&11&\textbf{1}&\textbf{6}\\
        L2V-Pho&\textbf{1}&\textbf{2}&11&\textbf{1}&\textbf{6}\\
        L2V-Inv&6&12&8&5&8\\
        L2V-Fam&\textbf{1}&\textbf{2}&11&\textbf{1}&\textbf{6}\\
        L2V-Geo&6&12&8&5&8\\
    \midrule
        \rowcolor[gray]{0.9}Max&1&1&1&1&1\\
    \bottomrule[1.2pt]
    \end{tabular}
    \caption{
    Ranking of the results from different methods of selecting source languages in MSLT, especially in scenarios where English is consistently incorporated into the set.
    The patterns are similar; Lang2Vec-based criteria demonstrate reasonable performance.
    }
    \label{tab:table2}
\end{table}

\subsection{Evaluation on Selection Criteria}
Table \ref{tab:table1} represents the results from experiments on source language selection criteria presented in \S\ref{section:5.3}.
The scores presented in the table are the averages of all scores for each task across target languages.
Target languages will have the same language set when the selection criterion applied is identical.

The findings suggest that selection criteria based on the statistical properties of language models, referenced in Sections \S\ref{sec:5.3.1} and \S\ref{sec:5.3.2}, do not substantially improve the performance of XLT.
In contrast, selecting source languages based on linguistic diversity demonstrates a significant performance gain. 
This implies that in MSLT, prioritizing diversity within the source language combination is more important than relying on the characteristics of corpora used in pretraining LMs.

Furthermore, although the `Embedding' method, which measures diversity directly based on embeddings from LMs, led to a fairly reasonable performance gain, measuring diversity using Lang2Vec vectors demonstrated the best outcomes. This further reinforces the importance of properly considering the inherent properties of languages in NLP.

\subsection{Language Selection Sticking to English}
\label{section:5.5}
This section features a case-study assuming that one of the source languages is necessarily English, a common scenario in practical applications.
Table \ref{tab:table2} reports that even in cases where English is always included in combinations, selecting other source languages based on diversity (or least similarity) according to Lang2Vec still exhibits reasonable performance.

\begin{figure*}[!t]
    \centering
    \includegraphics[width=\textwidth]{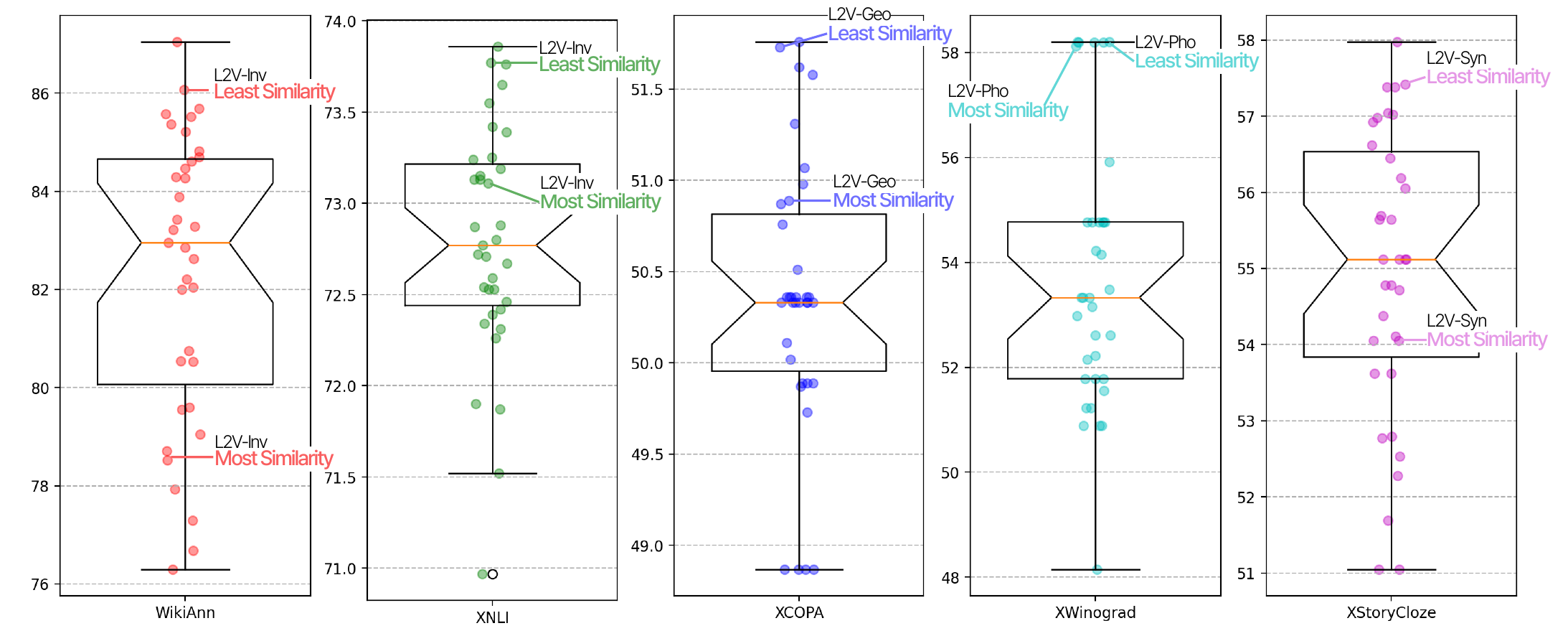}
    \caption{
    Visualization of performance for all (35) possible language sets. 
    We observe that language combinations with the \textbf{least} similarity (\textbf{high} diversity) yield better performance than those with the \textbf{most} similarity (\textbf{low} diversity).
    }
    \label{fig:figure6}
\end{figure*}

\section{Discussion}

\begin{table}[t]
    \small
    \centering
    \begin{tabular}{l c c c c c}
    \toprule[1.2pt]
        \textbf{Datasets} & \textbf{Top1} & \textbf{Top2} & \textbf{Top3} & \textbf{Top4} & \textbf{Top5}  \\
    \midrule
        \multirow{3}{*}{\shortstack[c]{\textbf{WikiAnn}}} & \cellcolor[HTML]{FCE38A}de & \cellcolor[HTML]{FCE38A}de & \cellcolor[HTML]{F38181}ar & \cellcolor[HTML]{F38181}ar & \cellcolor[HTML]{F38181}ar \\
        & \cellcolor[HTML]{94BDE0}fr & \cellcolor[HTML]{EAFFD0}es & \cellcolor[HTML]{FCE38A}de & \cellcolor[HTML]{EAFFD0}es & \cellcolor[HTML]{FCE38A}de \\
        & \cellcolor[HTML]{95E1D3}zh & \cellcolor[HTML]{95E1D3}zh & \cellcolor[HTML]{95E1D3}zh & \cellcolor[HTML]{95E1D3}zh & \cellcolor[HTML]{F2D1D1}ru \\
    \midrule
        \multirow{3}{*}{\shortstack[c]{\textbf{XNLI}}} & \cellcolor[HTML]{F38181}ar & \cellcolor[HTML]{FCE38A}de & \cellcolor[HTML]{F38181}ar & \cellcolor[HTML]{F38181}ar & \cellcolor[HTML]{F38181}ar \\
        & \cellcolor[HTML]{FCE38A}de & \cellcolor[HTML]{EAFFD0}es & \cellcolor[HTML]{FCE38A}de & \cellcolor[HTML]{F2D1D1}ru & \cellcolor[HTML]{94BDE0}fr \\
        & \cellcolor[HTML]{F2D1D1}ru & \cellcolor[HTML]{95E1D3}zh & \cellcolor[HTML]{95E1D3}zh & \cellcolor[HTML]{95E1D3}zh & \cellcolor[HTML]{95E1D3}zh \\
    \midrule
        \multirow{3}{*}{\shortstack[c]{\textbf{XCOPA}}} & \cellcolor[HTML]{FCE38A}de & \cellcolor[HTML]{F38181}ar & \cellcolor[HTML]{94BDE0}fr & \cellcolor[HTML]{F38181}ar & \cellcolor[HTML]{F38181}ar \\
        & en & \cellcolor[HTML]{EAFFD0}es & \cellcolor[HTML]{F2D1D1}ru & \cellcolor[HTML]{FCE38A}de & \cellcolor[HTML]{FCE38A}de \\
        & \cellcolor[HTML]{95E1D3}zh & \cellcolor[HTML]{95E1D3}zh & \cellcolor[HTML]{95E1D3}zh & \cellcolor[HTML]{95E1D3}zh & \cellcolor[HTML]{EAFFD0}es \\
    \midrule
        \multirow{3}{*}{\shortstack[c]{\textbf{XWinograd}}} & \cellcolor[HTML]{F38181}ar & \cellcolor[HTML]{FCE38A}de & \cellcolor[HTML]{FCE38A}de & en & \cellcolor[HTML]{EAFFD0}es \\
        & en & en & \cellcolor[HTML]{EAFFD0}es & \cellcolor[HTML]{EAFFD0}es & \cellcolor[HTML]{94BDE0}fr \\
        & \cellcolor[HTML]{95E1D3}zh & \cellcolor[HTML]{F2D1D1}ru & \cellcolor[HTML]{95E1D3}zh & \cellcolor[HTML]{F2D1D1}ru & \cellcolor[HTML]{F2D1D1}ru \\
    \midrule
        \multirow{3}{*}{\shortstack[c]{\textbf{XStoryCloze}}} & \cellcolor[HTML]{F38181}ar & \cellcolor[HTML]{F38181}ar & \cellcolor[HTML]{F38181}ar & \cellcolor[HTML]{EAFFD0}es & \cellcolor[HTML]{F38181}ar \\
        & \cellcolor[HTML]{FCE38A}de & \cellcolor[HTML]{FCE38A}de & \cellcolor[HTML]{FCE38A}de & \cellcolor[HTML]{94BDE0}fr & \cellcolor[HTML]{94BDE0}fr \\
        & \cellcolor[HTML]{F2D1D1}ru & \cellcolor[HTML]{95E1D3}zh & \cellcolor[HTML]{EAFFD0}es & \cellcolor[HTML]{95E1D3}zh & \cellcolor[HTML]{95E1D3}zh \\
    \bottomrule[1.2pt]
    \end{tabular}
    \caption{The five most effective combinations of source languages for five specific tasks. 
    }
    \label{tab:table3}
\end{table}

\subsection{Which Language is Suitable for MSLT?}
\label{section:6.1}
To identify the necessary conditions of the languages comprising the optimal set, we manually examine top-ranked language combinations for each task in practice.
Table \ref{tab:table3} showcases the top five language combinations that exhibit the best performance for each task.
The results reveal that, across all tasks, Chinese (\texttt{zh}), Arabic (\texttt{ar}), and German (\texttt{de}) are the most commonly used languages, with frequencies of 17, 15, and 15, respectively.
Furthermore, we find that each of the top 5 combinations includes two or more distinct writing systems, such as Arabic scripts for Arabic, Latin scripts for European languages, Chinese characters for Chinese, and Cyrillic scripts for Russian. 
This suggests that the diversity of writing systems can act as a key factor in constructing good language sets for MSLT.

To further investigate this perspective, we conduct an extra experiment whose outcomes are reported in Table \ref{tab:table4}.
In this experiment, we categorize source language sets (in total, 35) into three types based on the extent of diversity in their writing systems, and then compare the performance of each group.
The outcomes show that combinations of language with distinct writing systems always outperform other cases, implying the importance of diversity in writing systems among languages.
We believe that Lang2Vec-based criteria can properly exploit similar linguistic information.

\begin{table}[t]
    \footnotesize
    \centering
    \renewcommand{\arraystretch}{0.9}
    \begin{tabular}{l c c c}
        \toprule[1.2pt]
         \multirow{2.5}{*}{\shortstack[c]{\textbf{Dataset}}} & Case 1 & Case 2 & Case 3 \\
        \cmidrule{2-4}
         & $a=b=c$ & $a=b\ne c$ & $a\ne b\ne c$ \\
         \midrule
         WikiAnn & 72.26 & 72.69 & \textbf{73.07}\\ 
         XNLI & 80.36 & 81.54 & \textbf{84.02}\\
         XCOPA & 52.82 & 53.06 & \textbf{53.10}\\
         XWinograd & 52.33 & 52.89 & \textbf{53.40}\\
         XStoryCloze & 52.54 & 53.32 & \textbf{53.62}\\
        \bottomrule[1.2pt]
    \end{tabular}
    \caption{Test accuracy results based on the diversity of writing systems. 
    Writing system diversity in language combinations is categorized into three levels: Case 1, with all three languages sharing the same system; Case 2, with two languages sharing a system; and Case 3, with each language having a different system. 
    The outcomes show that combinations of language with distinct writing systems always outperform other cases.
    }
    \label{tab:table4}
\end{table}

\subsection{Is Increasing Diversity Really Helpful?}
\label{section:6.2}
In \S\ref{sec:5.3.3}, we follow the intuition that selecting language combinations with elements exhibiting distinct characteristics is beneficial.
In fact, we can also consider the opposite direction, grouping languages with relatively high similarities rather than low ones.
To determine which direction is more helpful for MSLT, we conduct an experiment using the two contrastive approaches.

Experimental results in Figure \ref{fig:figure6} reveal that constructing combinations with maximum diversity (as in \S\ref{sec:5.3.3}) yields better performance than combinations having minimum diversity (i.e., high similarity) across the majority of tasks.\footnote{Full experimental results are listed in Appendix, Table \ref{appendix:table6}.}
Notably, there is a significant performance gap between the two contrasting strategies for WikiAnn. 
We also observe similar patterns for other datasets, confirming the desirability of our decision to consider the diversity of source languages in \S\ref{sec:5.3.3}.

\section{Conclusion}
This work demonstrates the positive impact of Multi-Source Language Training (MSLT) in cross-lingual transfer (XLT). By leveraging multiple source languages, MSLT facilitates the learning of language-agnostic features in multilingual LMs.
We provide qualitative and quantitative evidence suggesting that this approach promotes cross-lingual semantic alignment, leading to better XLT performance compared to using a single source language.
Furthermore, we identify the importance of selecting the optimal size and composition of source language sets for MSLT. 
To guide this selection efficiently, we propose diverse criteria based on the statistical and linguistic properties of candidate languages, such as Lang2Vec.
Importantly, MSLT demonstrates effectiveness across various configurations, including different tasks, architectures, and training paradigms.
We hope this study serves as a springboard for further exploration of factors influencing the development of effective cross-lingual transfer methodologies.

\section*{Limitations}
\paragraph{Limited source language pools}
Given the practical availability of language-specific datasets and the ease of experimentation, this paper designates the top 7 languages that appeared most frequently in various multilingual datasets as source language candidates. By analyzing the permutations of selecting three out of these 7 languages, we aim to provide insights based on a wide range of scenarios to establish relevance in a generalized setting. However, it is important to note that further experiments with a greater number of languages as candidates for source languages are left as future work.

\paragraph{Comparison with other source language selection methods} 
Our study shares many similarities with other research on source language selection for cross-lingual transfer, aiming to find the optimal source languages. However, many selection methods tend to focus on the correlation between the source and target languages, making the selection process target-specific and often favoring single-language choices, which makes it challenging to quantitatively compare them with our approach.

\section*{Ethics Statement}
In this study, we utilize models and datasets publicly available on Huggingface, ensuring that no ethical issues are associated with their usage. Additionally, we endeavor to minimize content related to social biases. 
However, we acknowledge the possibility that some biases may have been infused during the pretraining process of language models, highlighting the need for sufficient consideration in this regard.

\section*{Acknowledgements}
This work was supported by Institute of Information \& communications Technology Planning \& Evaluation (IITP) grant funded by the Korea government(MSIT) (No.RS-2020-II201373, Artificial Intelligence Graduate School Program(Hanyang University)).
This work was supported by Institute of Information \& communications Technology Planning \& Evaluation (IITP) under the artificial intelligence semiconductor support program to nurture the best talents (IITP-2024-RS-2023-00253914) grant funded by the Korea government(MSIT).
This work was supported by the National Research Foundation of Korea(NRF) grant funded by the Korea government(*MSIT) (No.2018R1A5A7059549). *Ministry of Science and ICT.

\bibliography{custom}
\clearpage
\appendix

\section*{Appendix}
\label{sec:appendix}

\section{Dataset}
\label{appendix:dataset}
\textbf{WikiAnn} is a named entity recognition dataset sourced from Wikipedia, comprising 282 language data segments.
In our experimental setup, we utilize 7 languages (Arabic (\texttt{ar}), German (\texttt{de}), English (\texttt{en}), Spanish (\texttt{es}), French (\texttt{fr}), Russian (\texttt{ru}), Chinese (\texttt{zh})) as the source language and 8 languages (Indonesian (\texttt{id}), Greek (\texttt{el}), Hebrew (\texttt{he}), Finnish (\texttt{fi}), Thai (\texttt{th}), Turkish (\texttt{tr}), Japanese (\texttt{ja}), Korean (\texttt{ko})) as the target language for cross-lingual transfer.

\hfill \break
\textbf{PAWS-X} comprises 23,659 human-translated evaluation pairs and 296,406 machine-translated training pairs across six typologically distinct languages: French (\texttt{fr}), Spanish (\texttt{es}), German (\texttt{de}), Chinese (\texttt{zh}), Japanese (\texttt{ja}), and Korean (\texttt{ko}). These translated pairs originate from examples within PAWS-Wiki.
Two languages that are not included in the source language are used as target languages.

\hfill \break
\textbf{XNLI} is an evaluation corpus for language transfer and cross-lingual sentence classification in 15 languages.
XNLI cover 15 languages: Arabic (\texttt{ar}), Bulgarian (\texttt{bg}), Chinese (\texttt{zh}), English (\texttt{en}), French (\texttt{fr}), Greek (\texttt{el}), German (\texttt{de}), Hindi (\texttt{hi}), Russian (\texttt{ru}), Spanish (\texttt{es}), Swahili (\texttt{sw}), Thai (\texttt{th}), Turkish (\texttt{tr}), Urdu (\texttt{ur}), and Vietnamese (\texttt{vi}).
Among the aforementioned 15 languages, all except the one employed as the source language in our experiment serve as target languages.

\hfill \break
\textbf{XCOPA}, known as Cross-lingual Choice of Plausible Alternatives, is a benchmark designed to evaluate machine learning models' capability to transfer commonsense reasoning across different languages. This dataset comprises translations and re-annotations of the English COPA, encompassing 11 languages (Estonian (\texttt{et}), Haitian (\texttt{ht}), Indonesian (\texttt{id}), Italian (\texttt{it}), Quechua (\texttt{qu}), Swahili (\texttt{sw}), Tamil (\texttt{ta}), Thai (\texttt{th}), Turkish (\texttt{tr})) from diverse language families and regions worldwide.
All languages in XCOPA are evaluated as target languages by instruction-tuned models.

\hfill \break
\textbf{XWinograd} is a multilingual benchmark designed for assessing commonsense reasoning, consisting of Winograd Schema Challenge problems presented in six languages. The task entails selecting the most plausible sentence from slightly varied options.
We used Japanese (\texttt{ja}) and Portuguese (\texttt{pt}) as target languages for our experiments.

\hfill \break
\textbf{XStoryCloze}
comprises the English StoryCloze dataset professionally translated into 10 non-English languages. 
Out of these, we chose six languages as target languages, ensuring they did not overlap with the source language set.
The selected target languages are as follows: Basque(\texttt{eu}), Hindi (\texttt{hi}), Indonesian (\texttt{id}), Burmese (\texttt{my}), Swahili (\texttt{sw}), and Telugu (\texttt{te}).

\section{Training Details}
\label{appendix:hyperparameters}
We report the hyperparameters used for training each model in Table \ref{appendix:table5}. 
For XLM-R$_{\text{Base}}$ and XLM-R$_{\text{Large}}$, we conducted a hyperparameter search to find an appropriate learning rate for training. As for BLOOM-7B, we attempted to follow the settings from the original LoRA-tuned model,\footnote{\url{https://huggingface.co/MBZUAI/bactrian-x-bloom-7b1-lora}} but due to the lack of computation resources, we were unable to replicate the original settings such as batch size 128. Instead, we adjusted the training steps the same as the original LoRA-tuning setting. For the BLOOM model, prompt strategies employed for instruction-tuning and inference are listed in Table \ref{appendix:table7} and Table \ref{appendix:table8}, respectively.

\begin{table}[h]
    \centering
    \small
    \renewcommand{\arraystretch}{1.2}
    \setlength{\tabcolsep}{3pt}
    \begin{tabular}{l c c c}
        \toprule[1.2pt]
        \textbf{Model} & XLM-R$_\text{B}$ & XLM-R$_\text{L}$ & BLOOM$_\text{7B}$ \\
        \midrule
        Common & \multicolumn{3}{c}{cutoff\_length: 512 \& weight decay: 0.01}\\
        \midrule
        Learning rate & 2e-5 & 5e-6 & 3e-4 \\
        LoRA$_r$ & - & - & 64 \\
        LoRA$_m$ & - & - & q,k,v \\
        \bottomrule[0.9pt]
    \end{tabular}
    
    \setlength{\tabcolsep}{6pt}
    \begin{tabular}{l c c c c}
        \toprule[0.9pt]
         \textbf{Task} & WikiAnn & XNLI & PAWS-X & Bactrian-X \\
         \midrule
         Batch & 16 & 16 & 16 & 4 \\
         Step & 243K & 12.5K & 30K & 5K \\
         GA & 1 & 1 & 1 & 3 \\
         \bottomrule[1.2pt]
    \end{tabular}
    \caption{Best hyperparameters for each task and model. LoRA$_r$ indicates the rank of LoRA, LoRA$_m$ represents the position of LoRA's module, and GA stands for Gradient Accumulation.}
    \label{appendix:table5}
\end{table}

\begin{table*}[t]
    \centering
    \small
    \begin{tabular}{l c c c c c c c c c c c c c c}
    \toprule[1.2pt]
        \multirow{2.5}{*}{\shortstack[c]{Method}}& \multicolumn{2}{c}{WikiAnn} && \multicolumn{2}{c}{XNLI}&& \multicolumn{2}{c}{XCOPA} && \multicolumn{2}{c}{XWinograd}&& \multicolumn{2}{c}{XStoryCloze} \\
        \cmidrule{2-3}\cmidrule{5-6}\cmidrule{8-9}\cmidrule{11-12}\cmidrule{14-15}
         & Most & Least && Most & Least && Most & Least && Most & Least && Most & Least \\
         \midrule
        Embedding&77.93&85.58&&72.88&72.26&&48.85&48.85&&52.61&53.33&&55.12&56.98 \\
        L2V-Syn&78.52&85.69&&72.87&73.76&&50.32&51.57&&\textbf{58.21}&55.92&&54.05&\textbf{57.42} \\
        L2V-Pho&76.68&84.82&&72.88&73.39&&49.88&49.88&&58.20&\textbf{58.21}&&51.05&55.12 \\
        L2V-Inv&78.71&\textbf{86.07}&&73.13&\textbf{73.77}&&50.85&50.50&&52.98&58.20&&56.92&56.45 \\
        L2V-Fam&82.95&84.82&&71.90&73.39&&50.75&49.88&&58.20&\textbf{58.21}&&53.62&55.12 \\
        L2V-Geo&78.71&85.58&&73.13&73.24&&50.85&\textbf{51.72}&&52.98&58.12&&56.92&56.62 \\
    \midrule
        AVG&78.92&85.43&&72.80&73.30&&50.25&50.40&&55.53&57.00&&54.61&56.29 \\
    \bottomrule
    \end{tabular}
    \caption{Performance comparison of language combinations with high and low inter-lingual similarity. Accuracy is used as the evaluation metric for all tasks. The best results for each task are in \textbf{bold}.}
    \label{appendix:table6}
\end{table*}

\newcolumntype{T}{>{\raggedright}p{4cm}}
\newcolumntype{X}{>{\raggedright}p{11cm}}
\begin{table*}[t]
    \centering
    \begin{tabular}{T X}
        \toprule[1.2pt]
         \textbf{Only Instruction}: & \#\#\# Input: \cr
         & What is the capital of France? \cr
         &\cr
         & \#\#\# Output: \cr
         & The capital city of France is Paris.\cr
         \midrule
         \textbf{Instruction with input}: & \#\#\# Input: \cr
         & Evaluate this sentence for spelling and grammar mistakes. \cr
         & He finished his meal and left the restaurant. \cr
         &\cr
         & \#\#\# Output: \cr
         & There are two spelling errors in the sentence. The corrected sentence should be: "He finished his meal and left the restaurant."\cr
    \bottomrule[1.2pt]
    \end{tabular}
    \caption{Prompting strategies for instruction tuning.}
    \label{appendix:table7}
\end{table*}

\begin{table*}[t]
    \centering
    \begin{tabular}{T X}
        \toprule[1.2pt]
        \textbf{XCOPA} & \#\#\# Input: \cr
        & Answer the number of options which is more plausible for the effect of the situation "Mees keeras kraani lahti." \cr
        &\cr
        &1. Tualett täitus veega.\cr
        &2. Tilast voolas vett.\cr
        & \cr
        & \#\#\# Output: \cr
        \midrule
        \textbf{XWinograd} & \#\#\# Input: \cr
        & Answer the number of option which is more plausible for the blank.\cr
        & \begin{CJK}{UTF8}{min}"チンパンジーはリナックスを使えなかった。\_が動物だからだ。"\end{CJK} \cr
        & \cr
        & 1. \begin{CJK}{UTF8}{min}チンパンジー\end{CJK} \cr
        & 2. \begin{CJK}{UTF8}{min}リナックス\end{CJK} \cr
        & \cr
        & \#\#\# Output: \cr
        \midrule
        \textbf{XStoryCloze} & \#\#\# Input: \cr
        & Answer the number of option which is more natural for the end of story.
"Rick tumbuh di keluarga bermasalah. Dia tidak pernah menerima dukungan dari keluarga, dan menjadi anggota geng. Tak butuh waktu lama sampai Rick tertembak dalam sebuah perampokan. Peristiwa itu membuatnya insaf." \cr
        & \cr
        & 1. Kini dia bahagia.\cr
        & 2. Dia menjadi anggota geng.\cr
        & \cr
        & \#\#\# Output: \cr
        \bottomrule[1.2pt]
    \end{tabular}
    \caption{Prompting strategies for inference.}
    \label{appendix:table8}
\end{table*}

\end{document}